\documentclass[runningheads]{llncs}
\usepackage{graphicx}
\usepackage{amsmath}
\usepackage{booktabs}
\usepackage{hyperref}
\usepackage{multirow}
\usepackage{enumerate}
\usepackage{ragged2e}
\usepackage{balance}
\usepackage{setspace}
\usepackage{amsfonts}
\title{Learning Better Sentence Representation with Syntax Information}
\author{Chen Yang}
\institute{University of Science and Technology of China}
\begin{document}
\maketitle 

\begin{abstract}
Sentence semantic understanding is a key topic in the field of natural language processing. Recently, contextualized word representations derived from pre-trained language models such as ELMO and BERT have shown significant improvements for a wide range of semantic tasks, e.g. question answering, text classification and sentiment analysis. However, how to add external knowledge to further improve the semantic modeling capability of model is worth probing. In this paper, we propose a novel approach to combining syntax information with a pre-trained language model. In order to evaluate the effect of the pre-training model, first, we introduce RNN-based and Transformer-based pre-trained language models; secondly, to better integrate external knowledge, such as syntactic information integrate with the pre-training model, we propose a dependency syntax expansion (DSE) model. For evaluation, we have selected two subtasks: sentence completion task and biological relation extraction task. The experimental results show that our model achieves 91.2\% accuracy, outperforming the baseline model by 37.8\% on sentence completion task. And it also gets competitive performance by 75.1\% $F_{1}$ score on relation extraction task.

\end{abstract}
\section{Introduction}
Over the past year, pre-training language models have attracted the attention of many natural language processing researchers. The current research work focused on pre-trained language models can be divided into three main categories: (1)applying contextualized word representations to downstream tasks\cite{sentiment_bert,passage_bert}; (2)probing what contextualized word representations learned from large scale corpus \ cite {tenney2019you, liu20linguistic}; (3)using external information to improve word representations\cite{ERNIE, ERNIE_baidu}. As further exploration of category (3), in this paper, we proposed a dependency syntax expansion (DSE) model to combine syntax information and pre-trained LMs. To be more specific, we propose a simple yet effective approach to use dependency syntax information by a triple form. Such triple form can be directly integrated with the output of RNN-based or Transformer-based pre-trained language model. Compared with other dependency parser tree based methods \cite{tree_lstm,tree_ddi,hierarchical_ddi}, the proposed DSE can be trained in a mini-batch way and is shown to be more effective.

In order to verify the effectiveness of the method, we choose two sentence-level tasks to evaluate our model. The first task is sentence completion, which requires the model to choose the most appropriate word or phrase from a given set to complete a sentence. Zweig et al.\cite{zweig2011microsoft} present a public standard dataset on this task, which is called the Microsoft Research Sentence Completion Challenge (MSCC). The dataset collects 1,024 sentences from \textit{Sherlock Holmes} novels and sentence has one word missing and five candidate words with similar occurrence statistics. Much study\cite{rnn_skip_gram,memory_rnn,} has shown that pre-trained language model and syntax information is beneficial on MSCC. The second task is a relation extraction task, which aims to classify two mentioned entities in a sentence into a specific type. Specifically, we select SemEval2013 DDI Extraction challenge data as our experimental dataset, which is a benchmark for biological relation extraction. It is notable that previous studies \cite{tree_ddi,hierarchical_ddi} have also proved that dependency syntax information plays an important role in relation extraction. 

Additionally, we present English Sentence Gap-Filling dataset (ESG), a new dataset for sentence completion task which is mined from standardized English examinations. Compared to  MSCC data, our data contains more question types and richer grammatical phenomena. Some examples are shown in Table \ref{tab:question_example}. More details about the ESG dataset can be seen in Section 4.1.

\begin{table}[htbp]
  \centering
\vspace{-0.4cm}
  \caption{The three types of question form are shown in the table from left to right are: single word,consecutive phrases and short sentences}
  \scalebox{0.75}{
    \begin{tabular}{l|l|l}
    \toprule
    1. \_\_\_  become taller, jack drinks & 2.The old man is \_\_\_ Japanese \_\_\_  & 3.From the passage,we can learn that \_\_\_\_ \\
    three boxes of milk every day. & Chinese.     & A. he will be nervous all the time \\
    A.so that    & A.either,or  & B. he will find himself nervous all the time \\
    B.in order not to & B.not only,but also & C. you will find him nervous all the time \\
    C.since      & C.both, and  & D.everyone will find him nervous all the  \\
    D.in order to & D.either, nor & time \\
    \bottomrule
    \end{tabular}}
 \label{tab:question_example}%
  \vspace{-0.4cm}
\end{table}%


The contributions of our paper are as follows:
\begin{itemize}
\item We propose a novel syntax expansion structure and combine it with an advanced pre-trained language model.
\item We conduct our dependency syntax expansion model on two sentence-level and design several groups of experiments to evaluate the effectiveness of our model.
\item We construct a complete experimental dataset of the English examinations automatic answering question, which can serve as a standard dataset for the sentence completion task.
\end{itemize}

\section{Related Work}

In pre-trained language models, ELMO and BERT (Peters et al. \cite{peters2018deep},Devin et al. \cite{devlin2018bert}) are two of the most highlight works in the past year. The former used BiLSTM and trained with a coupled language model objective. The latter employed bidirectional Transformer encoder (Vaswani et al.\cite{vaswani2017attention}) instead of BiLSTM and predict the masked words according to the left and the right context.  Subsequently, further studies from different perspectives were published based on BERT and ELMO. Sun et al. \cite{sentiment_bert} and Nogueira et al. \cite{passage_bert} applied pre-trained BERT model on two downstream tasks respectively: sentiment analysis and passage ranking, and both achieved state-of-the-art results. Liu et al. \cite{liu2019linguistic} and Tenney et al. \cite{tenney2019you}) proved that these word representations provided by ELMO or BERT have the ability to capture syntactic and semantic information.  Zhang et al. \cite{ERNIE} and Sun et al. \cite{ERNIE_baidu} considered incorporating knowledge graphs or entity information with contextualized embeddings to enhance language representations. Lee et al. \cite{lee2019biobert} and Beltagy et al. \cite{beltagy2019scibert} used specific domain data to fine-tune BERT, which can boost the performance on domain specific tasks.

In sentence completion task, Zweig et al. \cite{zweig2012computational} proposed two baseline models that are based on n-gram language model and latent semantic analysis, which get the accuracy of 39 \% and 49\% respectively. Mikolov et al. \cite{rnn_skip_gram}  combined skip-gram with RNN-based language model and obtained 58.9\% accuracy in MSCC. Park et al.\cite{word_rnn}  exceeded the previous best results by using a bidirectional word-level RNN . In addition to the transformation of the language model architecture, other studies considered combing syntactic information with models.  Piotr Mirowski et al. \cite{dep_rnn} integrated syntax information with RNN language model, which achieved an absolute improvement of 10 \% improvement compared with previous RNN-based baseline. Zhang et al. \cite{tree_lstm} used a tree-structured LSTM model to infuse dependency path information with model.

In DDI2013 Extraction task, Zhang et al \cite{hierarchical_ddi} intergrated the shortest dependent path information and sentence sequence via hierarchical RNN network, which is superior to other state-of-the-art methods. Miwa et al. \cite{tree_ddi} used a bidirectional Tree-LSTM to capture the substructure information in the dependency tree, which achieved 12.1\% and 5.7\% relative error reductions in F1- score.

As can be seen from previous works on the sentence completion task and relation extraction task, almost all the models that leveraged syntax information rely on tree structure. In our DSE model, we use the triple form to linearize tree structure and then we utilize expansion syntax information to get better sentence representations, which has generality on sentence completion task and relation extraction task.

\section{Methodology}
\begin{figure}
\setlength{\abovecaptionskip}{-0.1cm}
\setlength{\belowcaptionskip}{-0.4cm}
    \centering
    \vspace{-0.4cm}
    \includegraphics[width=0.95\textwidth]{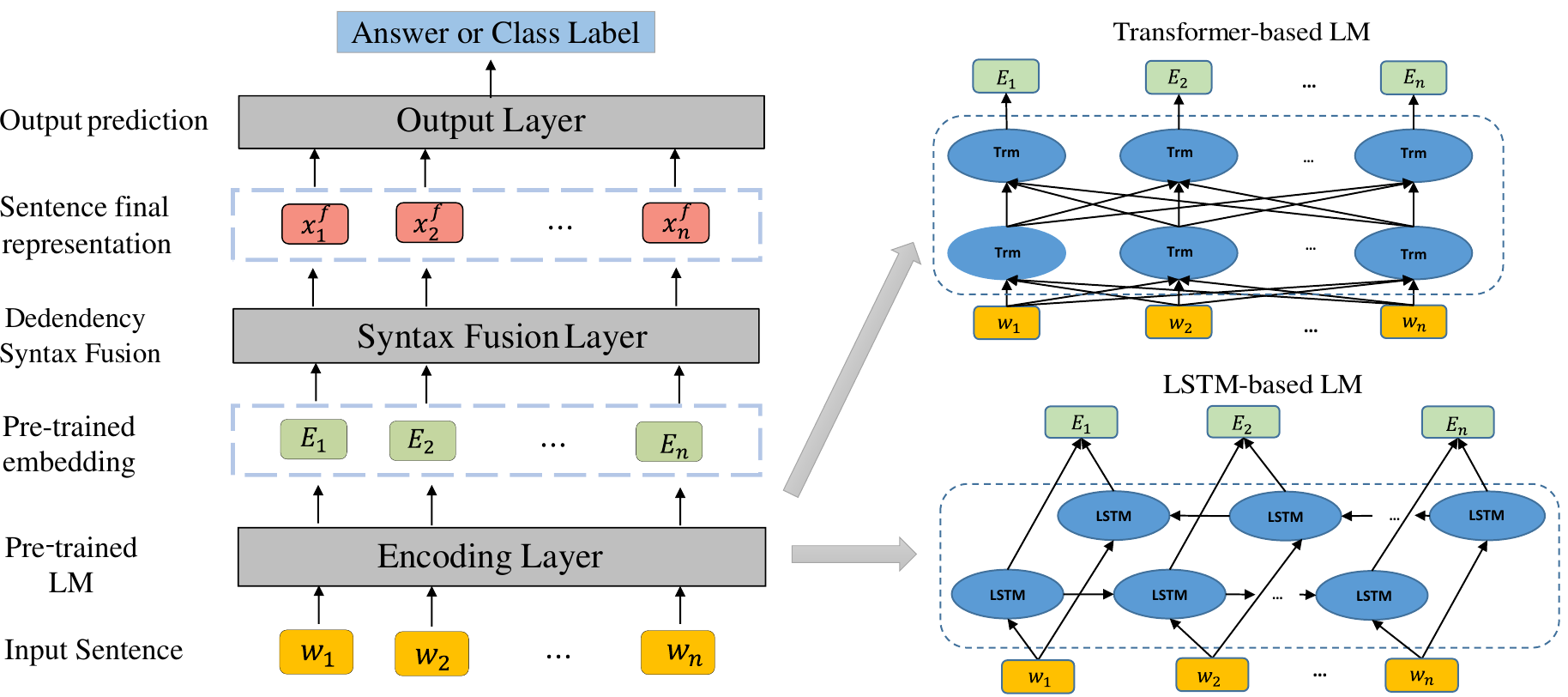}
    \caption{Architecture of our DSE model.}
    \label{fig:model}
     \vspace{-0.1cm}
\end{figure}

In this section, we present our DSE model in detail. Figure\ref{fig:model} shows the architecture of the model. It contains three components: (1) Encoding Layer: utilize pre-training language model to obtain word embeddings; (2) Dependency Syntax Fusion Layer: combine syntax information with pre-trained word embeddings and use BiLSTM to generate the final sentence representations.(3)Output Layer: get the most likely answer option or label according to the final semantic representations of sentences. And in this paper, we denote \({S}=\left[{w}_{1},{w}_{2}, \cdots, {w}_{n}\right]\) as a sentence consisting of n words.

\subsection{Encoding Layer} 
Generally, the encoding layer encodes each word in the sentence into a fixed-length vector and pre-trained word embeddings are a commonly used component in the encoding layer. In this subsection, we first obtain word embeddings using a bidirectional LSTM network (Hochreiter and Schmidhuber \cite{hochreiter1997long}) to train with two unidirectional LM objectives on the large corpus, which is similar to ELMO. And we name it biLM.  Secondly, we use BERT pre-trained model based Transformer architecture to get a different version of contextualized word embeddings. To be more specific, for biLM, the forward pass assigns the probability of the sentence by modeling the probability of word $w_k$ given $\left(w_{1}, \dots, w_{k-1}\right)$. In contrast, the backward pass runs over the sentence in reverse. The objective function jointly maximizes the log likelihood of the forward and backward pass:
\begin{equation}
\vspace{-0.1cm}
\begin{array}{l}
p\left(w_{1}, w_{2}, \ldots, w_{n}\right)=\prod_{k=1}^{n} p\left(w_{k} | w_{1}, w_{2}, \ldots, w_{k-1}\right)\\
p\left(w_{1}, w_{2}, \ldots, w_{n}\right)=\prod_{k=1}^{n} p\left(w_{k} | w_{k+1}, w_{k+2}, \ldots, w_{n}\right)
\end{array}
\label{eq:biLM}
\end{equation}

\begin{equation}
\begin{array}{l}
{\sum_{k=1}^{n}\left(\log p\left(w_{k} | w_{1}, \ldots, w_{k-1} ; \mathop {\mathop \theta \limits^ \leftarrow  }\nolimits_{LSTM} \right)\right.}\\ 
{\quad+\log p\left(w_{k} | w_{k+1}, \ldots, w_{N} ;\mathop {\mathop \theta \limits^ \to  }\nolimits_{LSTM}\right))}
\end{array}
\label{eq:biLM_loss}
\vspace{-0.2cm}
\end{equation}
where $\mathop {\mathop \theta \limits^ \leftarrow  }\nolimits_{LSTM}$ and $\mathop {\mathop \theta \limits^ \to  }\nolimits_{LSTM}$ are the parameters of the forward and backward LMs respectively. 

For BERT pre-trained model, due to Transformer architecture is not effective in modeling traditional language model objective, it uses a slightly modified language modeling objective that masks some tokens randomly and then predicts the likelihood of those masked tokens by softmax layer. In the training phase, we update the parameters of BERT but freeze the parameters of biLM.

\subsection{Dependency Syntax Fusion Layer}
\begin{figure}
\centering
 \vspace{-0.4cm}
\setlength{\abovecaptionskip}{-0.1cm}
\setlength{\belowcaptionskip}{-0.4cm}
\includegraphics[width=0.85\textwidth,height=2cm,scale=0.9]{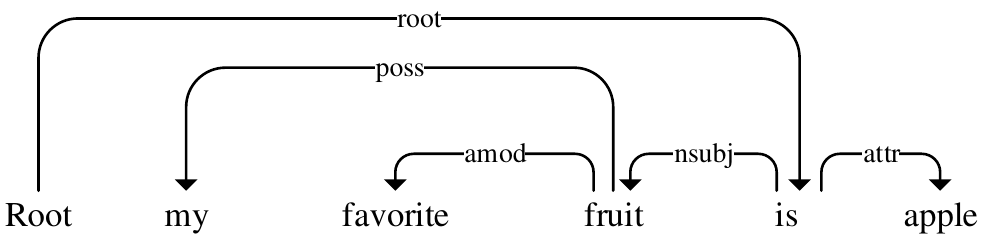}
\caption{Dependency tree for the sentence "my favorite fruit is apple."} \label{fig:dependency_tree}
 \vspace{-0.1cm}
\end{figure}

In this subsection, we describe how to combine explicit syntax information with pre-trained word embeddings. Dependency syntax information can reflect the directed grammatical relations between words in the sentence. In general, these dependencies are represented by binary trees. Figure \ref{fig:dependency_tree} shows the dependency analysis result of the sentence "\textit{my favorite fruit is apple}". We can see that each word in the sentence depends on a unique head. And much work has proved that these head-dependent relations are useful for many applications such as conference resolution, information extraction and question answering. Unfortunately, tree-based RNN models need to compute multiple times on different separate branches, which is time-consuming. Therefore, we propose a dependency tree expansion structure to alleviate this problem.
According to the unique correspondences mentioned above, we can represent head-dependent pairs and their relations by a triple form: (dependent, relation, head). For example, (fruit, nsubj, is) indicates that \textit{fruit} modifies \text{is} and the dependency relation of them is \textit{nsubj}. As illustrated in Figure~\ref{fig: unroll_tree}, we can convert the above dependency tree into an expanded tree by this approach.
\begin{figure}
    \centering
     \vspace{-0.3cm}
\setlength{\abovecaptionskip}{-0.1cm}
\setlength{\belowcaptionskip}{-0.4cm}
    \includegraphics[width=0.85\textwidth]{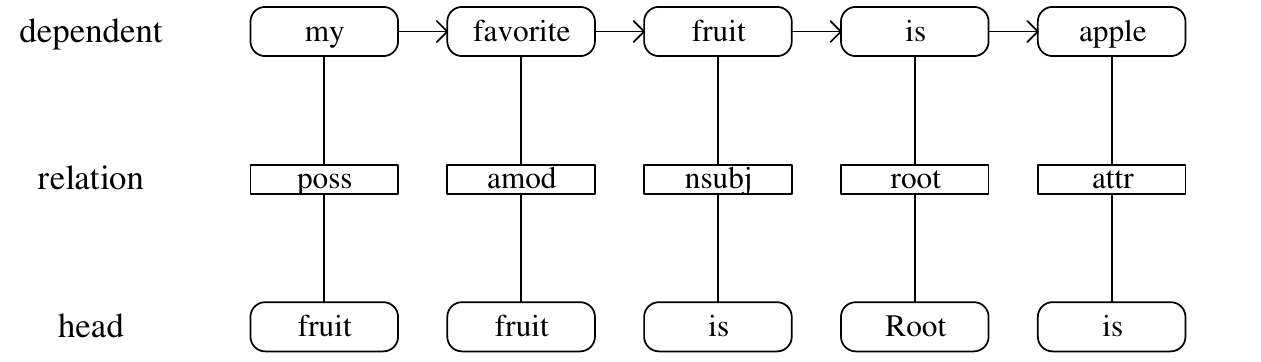}
    \caption{The structure of  dependency expansion tree}
    \label{fig:unroll_tree}
     \vspace{-0.1cm}
\end{figure}

Furthermore, by means of the triples form, we can easily integrate syntax information with contextualized word embeddings and generate new sentence representations. Specifically, we design two fusion functions to map a triple into a continuous vector in Equation~\ref{eq:fusion}
\begin{equation}
x_{i}=g\left(w_{i}, r_{i}, w_{i}^{H}\right)
\label{eq:fusion}
\vspace{-0.1cm}
\end{equation}
where $w_{i} $ , \(w_{i}^{H} \in \mathbb{R}^{d}\) , \(w_{i}^{H} \) is the head of \(w_{i}\) and \(d\) is the dimension of word embedding. $r_i$ is a low dimensional vector to indicate different dependency relation, which is learned as a parameter by the model. The fusion function \(g(\cdot)\) has two implementation methods by concatenating vectors and using gate mechanism to control the information flow respectively:
\begin{equation}
g\left(w_{i}, r_{i}, w_{i}^{H}\right)=w_{i} \oplus r_{i} \oplus w_{i}^{H}
\label{eq:concat}
\vspace{-0.1cm}
\end{equation}
\begin{equation}
\vspace{-0.15cm}
g\left(w_{i}, r_{i}, w_{i}^{H}\right)=w_{i}+\sigma\left(r_{i}\right) \otimes w_{i}^{H}
\label{eq:gate}
\end{equation}
where $\oplus$ represents concatenation operation, $\otimes$ represents cross product operation between two vectors and $\sigma$ means a sigmoid function. We use ${S}^{p} = [{x_1},{x_2},\ldots, {x_n}]$ to indicate the processed sentential sequence by fusion function.

Next, ${S}^{p}$ is entered into a BiLSTM network as the data input. The following formulas show how to calculate each gate and memory cell unit of a typical BiLSTM network:
\begin{equation}
\centering
\setlength{\abovedisplayskip}{0.2cm}
\setlength{\belowdisplayskip}{0.2cm}
\label{eq:9}
\begin{array}{l}
h_{i} = \left[\overrightarrow{h_{t}} \oplus \overleftarrow{h_{t}}\right]\\
h_{t} =o_{t} \tanh \left(c_{t}\right) \\
f_{t} =\sigma\left(W_{f} \cdot\left[C_{t-1}, h_{t-1}, x_{t}\right]+b_{f}\right)\\
i_{t} =\sigma\left(W_{i} \cdot\left[C_{t-1}, h_{t-1}, x_{t}\right]+b_{i}\right)\\
g_{t} =\tanh \left(W_{x c} \cdot\left[C_{t-1}, h_{t-1}, x_{t}\right]+b_{c}\right) \\
c_{t} =i_{t} g_{t}+f_{t} c_{t-1} \\ 
o_{t} =\sigma\left(W_{o} \cdot\left[C_{t}, h_{t-1}, x_{t}\right]+b_{o}\right) \\
\end{array}
\end{equation}
The final sentence representation is composed of the forward and backward output in the last position and it is defined as $S^{f}=\left[{x}_{1}^{f},{x}_{2}^{f},\ldots,{x}_{n}^{f}\right]$. 
\subsection{Output Layer}
In the output layer, we use a fully connected feed-forward network to make the final prediction. However, there is a slight difference between sentence completion task and relation extraction task on the objective function. On sentence completion task, we use a pairwise approach to rank four answer options in the question. The loss function is defined as:

\begin{equation}
\centering
\setlength{\abovedisplayskip}{0.2cm}
\setlength{\belowdisplayskip}{0.2cm}
\operatorname{loss}=\sum_{i=1,2,3}^{N}\max (0,- {f}({S}_{true}^f)+{f}({S}_{false\_i}^f) +\operatorname{margin})
\label{eq:qa_loss}
\end{equation}

$f(\cdot)$ compute the scores that are assigned to each candidate sentence and choose the completion with the highest score. Its formula is $f(x)=\sigma\left(u^{T} x\right)$, where $u \in \mathbb{R}^{k}$ is a learnable parameter.$[S_{true}^{f},S_{false\_1}^{f},S_{false\_2}^{f},S_{false\_3}^{f}]$ corresponds with the semantic representations of the right sentence and three wrong sentences. And \textit{margin} is a custom threshold value to widen the score between ${S}_{true}^{f}$ and ${S}_{false\_i}^{f}$ . 

On the DDI extraction task, we use negative log-likelihood in equation \ref{eq:ddi_loss} as loss function.
\begin{equation}
\centering
\setlength{\abovedisplayskip}{0.15cm}
\setlength{\belowdisplayskip}{0.15cm}
\operatorname{loss}= -\log p\left({y}_{i} | S_{{p_i}}^f\right) = -\log \left(\frac{\exp ({y}_{i})}{\sum_{j}^{5} \exp ({y}_{j})}\right)
\label{eq:ddi_loss}
\end{equation}
where $S_{{p_i}}^f$ is an entity pair in the sentence $S^{f}$ and  ${y}_{i}$ is the relation class of ${p}_{i}$. The probability of an entity pair $S_{{p_i}}^f$ belongs to class ${y}_{i}$ is calculated by softmax function. 

\section{Experiments}

\subsection{Dataset}
\noindent\textbf{ESG dataset} The resources of our ESG data are mainly from high school English multiple choice questions on English learning websites. And we preprocess the original dataset before the training phase as follows. Firstly, given that many emphasis question types will be repeatedly examined in different examination papers, we utilize editing distance algorithm to filter similar questions while the distance between questions is below the threshold of 8. Secondly, we use Stanford Parser to implement part-of-speech tagging and dependency parsing on our data.  Finally, 62,834 questions are saved and divided into training sets and test sets according to the ratio of approximately 9:1.


We classify the question types of data sets by part-of-speech and stemming information. The distribution of question types is shown in Figure 3. It demonstrated that ESG dataset contains a wide range of grammatical phenomena.

\begin{figure}
     \centering
     \vspace{-0.4cm}
\setlength{\abovecaptionskip}{-0.1cm}
\setlength{\belowcaptionskip}{-0.4cm}
     \includegraphics[width=0.8\textwidth]{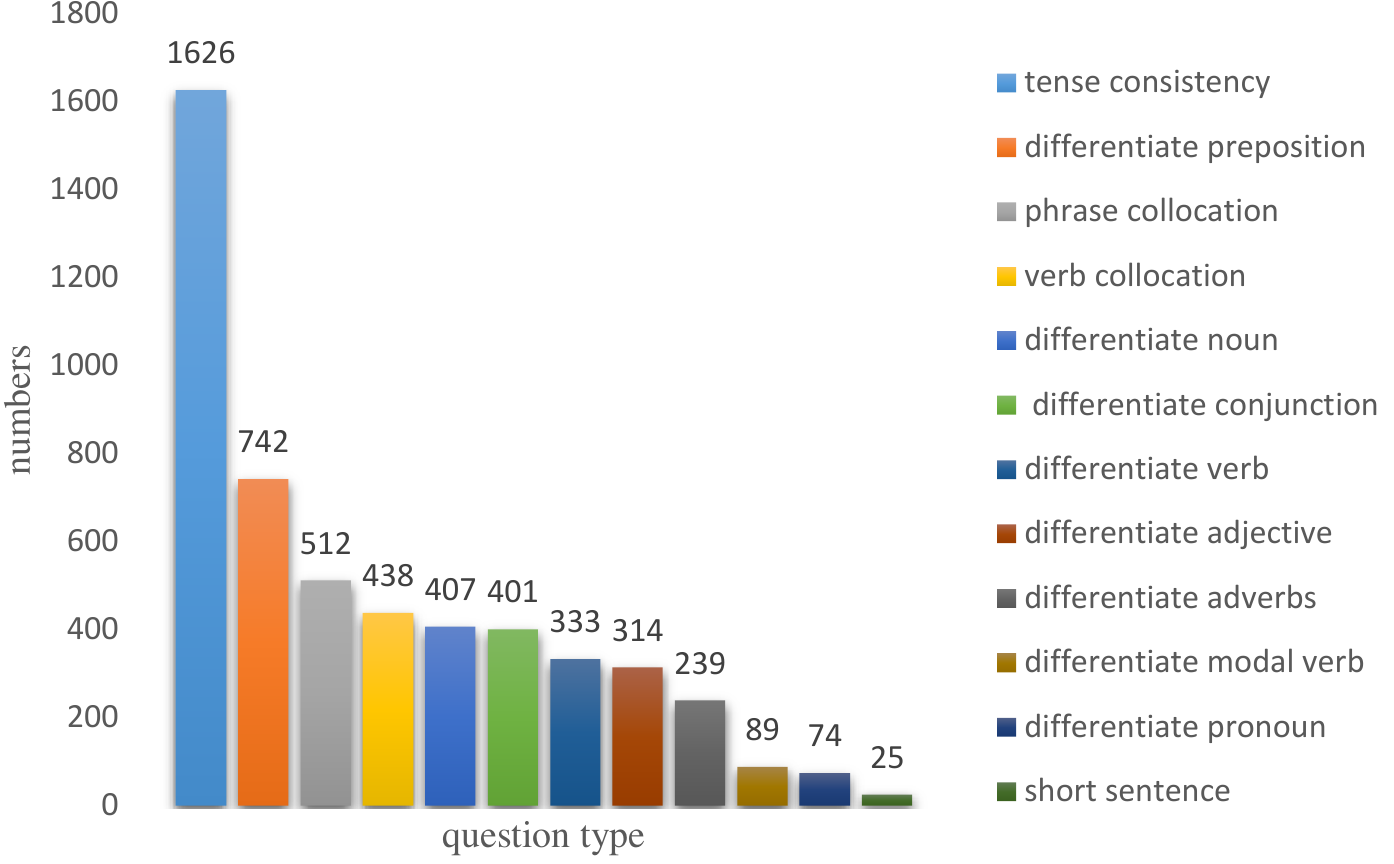}
     \caption{The distribution of question types}
     \label{fig:dataset}
     \vspace{-0.4cm}
\end{figure}

\noindent\textbf{DDI dataset} DDIExtraction 2013 database (https://www .cs.york.ac.uk/semeval-2013/task9/) comes from the DrugBank database and DrugMedline database. The main purpose of this task is to classify drug-drug interaction relation correctly from five categories including advice, effect, mechanism, int and negative. A detailed description of data can be seen in Table \ref{tab:ddi_data}.

\begin{table*}[ht]
  \centering
  \setlength{\abovecaptionskip}{-0.4cm}
\setlength{\belowcaptionskip}{0.1cm}
  \caption{The statistics of DDIExtraction 2013 datasets. The \emph{negative} means there is no any relation between two drug entities.}
    \begin{tabular}{ccccccc}
    \toprule
    \multirow{2}[4]{*}{} & \multicolumn{3}{c}{Train} & \multicolumn{3}{c}{Test} \\
    \midrule
          & DrugBank & MedLine & Overall & DrugBank & MedLine& Overall \\ 
\hline
    Positive & 3767  & 231   & 3998  & 884   & 92    & 976 \\
    Negative & 14445 & 1179  & 15624 & 2819  & 243   & 3062 \\
    Advice & 815   & 7     & 822   & 214   & 7     & 221 \\
    Effect & 1517  & 152   & 1669  & 298   & 62    & 360 \\
    Mechanism & 1257  & 62    & 1319  & 278   & 21    & 299 \\
    Int   & 178   & 10    & 188   & 94    & 2     & 96 \\
    \bottomrule
    \end{tabular}%
  \label{tab:ddi_data}%
   \vspace{-0.5cm}
\end{table*}%

\subsection{Experiments Detail}
In this subsection, we introduce the details of experiments on sentence completion task and DDI Extraction task. 

In sentence completion task, we use the PyTorch implementation of BERT with the pre-trained models supplied by Google(https://github.com/huggingface /
pytorch-pretrained-BERT). Considering BERT uses BPE as the tokenizer, which will split certain words into multiple sub-words, we use first word piece embedding as word embedding to ensure the sequence length of output is consistent with the original sentence length. We use the \textit{bert-base-cased} model ($\rm BERT_{base\_uncased}$) and pick 128 as max sequence length, 16 as the mini-batch size. We use BertAdam as the optimizer, with a learning rate of $\alpha$ = $2e-5$  and \textit{warmup } = $0.05$. The relation vector size is set to 200 and the hidden size of BiLSTM is set to 768. We train the model using 6 epochs and evaluate the accuracy of the model on the test set after each epoch of training. Additionally, in order to train biLM model, we collect 26G unannotated text corpus from Wikipedia, Gigaword and English learning websites.

In DDI task, we choose BioBERT\cite{lee2019biobert} model to provide word embeddings, which is a domain-adaptive version of BERT that pre-trained on biomedical domain corpora. We pick 350 as sequence length and 32 as minibatch size. The parameters of learning rate and \textit{warmup}  are changed into $3e-5$ and $0.1$. The BiLSTM has a hidden size of 512. Other hyper-parameters are the same as the former task configuration. Moreover, we use "drug1" and "drug2" to replace two entities that need to be predicted respectively before putting the sentence into an encoding layer. And other entities in the sentence are replaced with "drug0". Previous work has shown that this approach allows models to better understand the semantic relationships between entities.

\subsection{Experiments Result and Analysis}

\noindent\textbf{Evaluation on Sentence Completion Task} Table~\ref{tab:qa_res} shows the experimental result of baseline and our DSE model on sentence completion task. Our baseline model just uses a BiLSTM layer to encode the sentence representations and the input word vectors are generated by random initialization. And on the table, biLM models and \textit{$\rm BERT_{base\_uncased}$} models are trained without syntax fusion layer. We use accuracy to evaluate our model. As can be seen in Table~\ref{tab:qa_res}, the best performing result is 91.2\%, provided by dependency syntax expansion model with BERT. And we can see that pre-trained language models have a great improvement on the performance of model. Meanwhile, the DSE model with BERT is significantly better than the DSE model with biLM.  Adding syntactic information on the basis of pre-trained LM is effective, which can improve the accuracy(ACC) by 2 \% and  0.6 \% for biLM-based and BERT-based model respectively. Moreover, the advantages and disadvantages of two fusion methods (in equation~\ref{eq:concat} and equation~\ref{eq:gate}) are vary from model. For the biLM+DSE" model, the gate mechanism is better. But the conclusion is opposite to the $\rm BERT_{base\_uncased}$+DSE" model.

In addition, inspired by some works that fine-tune BERT by using specific domain data such as SciBERT\cite{beltagy2019scibert} and BioBERT\cite{lee2019biobert}, we extract additional 1.6G corpus for pre-training, which contains fine-grained words or phrases collected from error samples. We name this new version of BERT as $\rm BERT_{finetune}$. The comparison result between original BERT and $\rm BERT_{finetune}$ shows that fine-tune step can boost the performance of model stably on ESG dataset.
\noindent\textbf{DDI dataset} DDIExtraction 2013 database (https://www .cs.york.ac.uk/semeval-2013/task9/) comes from the DrugBank database and DrugMedline database. The main purpose of this task is to classify drug-drug interaction relation correctly from five categories including advice, effect, mechanism, int and negative. A detailed description of data can be seen in Table \ref{tab:ddi_data}.

\begin{table}[ht]
  \centering
  \vspace{-0.05cm}
  \setlength{\abovecaptionskip}{-0.5cm}
\setlength{\belowcaptionskip}{0.15cm}
  \caption{Sentence completion results on ESG dataset. 
  $(\cdot)$  indicates a specific fusion method in equation~\ref{eq:concat} or equation\ref{eq:gate}}
    \begin{tabular}{c|c}
    \toprule
    Model        & Acc(\%) \\
    \midrule
    \midrule
    baseline & 53.4 \\
        \midrule
    biLM    & 73.0 \\
    biLM+DSE(concat)  & 73.9 \\
    biLM+DSE(gate) & \textbf{75.9} \\
     \midrule
    $\rm BERT_{base\_uncased}$     & 90.3 \\
    $\rm BERT_{base\_uncased}$+DSE(gate)  & 90.7 \\
    $\rm BERT_{base\_uncased}$+DSE(concat) & 90.9 \\
   $\rm BERT_{finetune}$+DSE(concat)  &\textbf{91.2}\\
    \bottomrule
    \end{tabular}%
  \label{tab:qa_res}%
   \vspace{-0.5cm}
\end{table}%

Then, we construct a simple LSTM model based on our syntax expansion structure(Expansion-LSTM) and compare the training time between it and traditionally Tree-LSTM model in Table~\ref{tab:lstm_speed}. The result shows that our expanded tree structure is truly computationally cheap. 

\begin{figure}[t]
\vspace{-0.5cm}
\setlength{\abovecaptionskip}{-0.1cm}
\setlength{\belowcaptionskip}{-0.2cm}
    \centering
    \includegraphics[width=0.9\textwidth,height=5.5cm,scale=0.65]{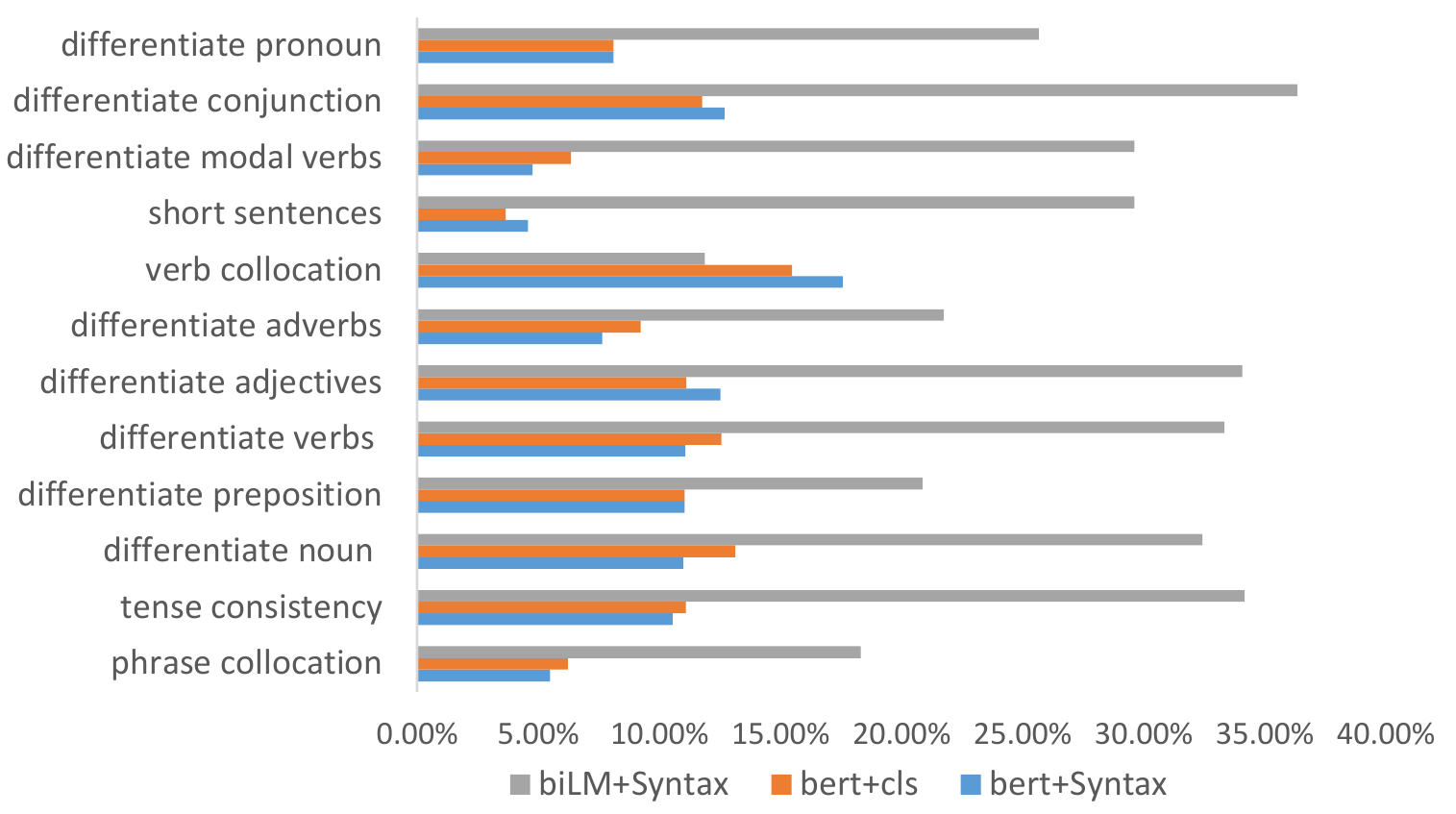}
    \caption{The proportion of error samples for different question types}
    \label{fig:error_ratio}
    \vspace{-0.2cm}
\end{figure}
\begin{table}[htb]
  \centering
  \setlength{\abovecaptionskip}{-0.3cm}
\setlength{\belowcaptionskip}{0.1cm}
  \caption{The comparison of model training time between Tree-LSTM and Expansion-LSTM.}
    \begin{tabular}{c|cc}
    \toprule
    Model        & Time (seconds per train epoch) \\
    \midrule
    \midrule
    Tree-LSTM    & 10431 \\
    Expansion-LSTM       & 2367 \\
    Expansion-LSTM +Tree-LSTM & 2587 \\
    \bottomrule
    \end{tabular}%
  \label{tab:lstm_speed}%
   \vspace{-0.5cm}
\end{table}%
\noindent\textbf{Error Analysis and Visualization on Sentence Completion Task} We analyze the proportion of error samples on different question types. From the results in Fig~\ref{fig:error_ratio}, we can see that the error rates in verbs differentiation, tenses consistency and phrase collocation have a significant decrease. It  benefits from syntax information and fine-tuned step. Furthermore, we use Bertviz, a visualization tool for explore the attention patterns of the pre-trained BERT model, to visual the difference of information learned by $\rm BERT_{base\_uncased}$ model and "$\rm BERT_{base\_uncased}$+DSE" model. 
\begin{figure}[htb]
\vspace{-0.5cm}
    \setlength{\abovecaptionskip}{-0.2cm}
\setlength{\belowcaptionskip}{-0.4cm}
    \includegraphics[width=0.9\textwidth,height=5.5cm,scale=0.7]{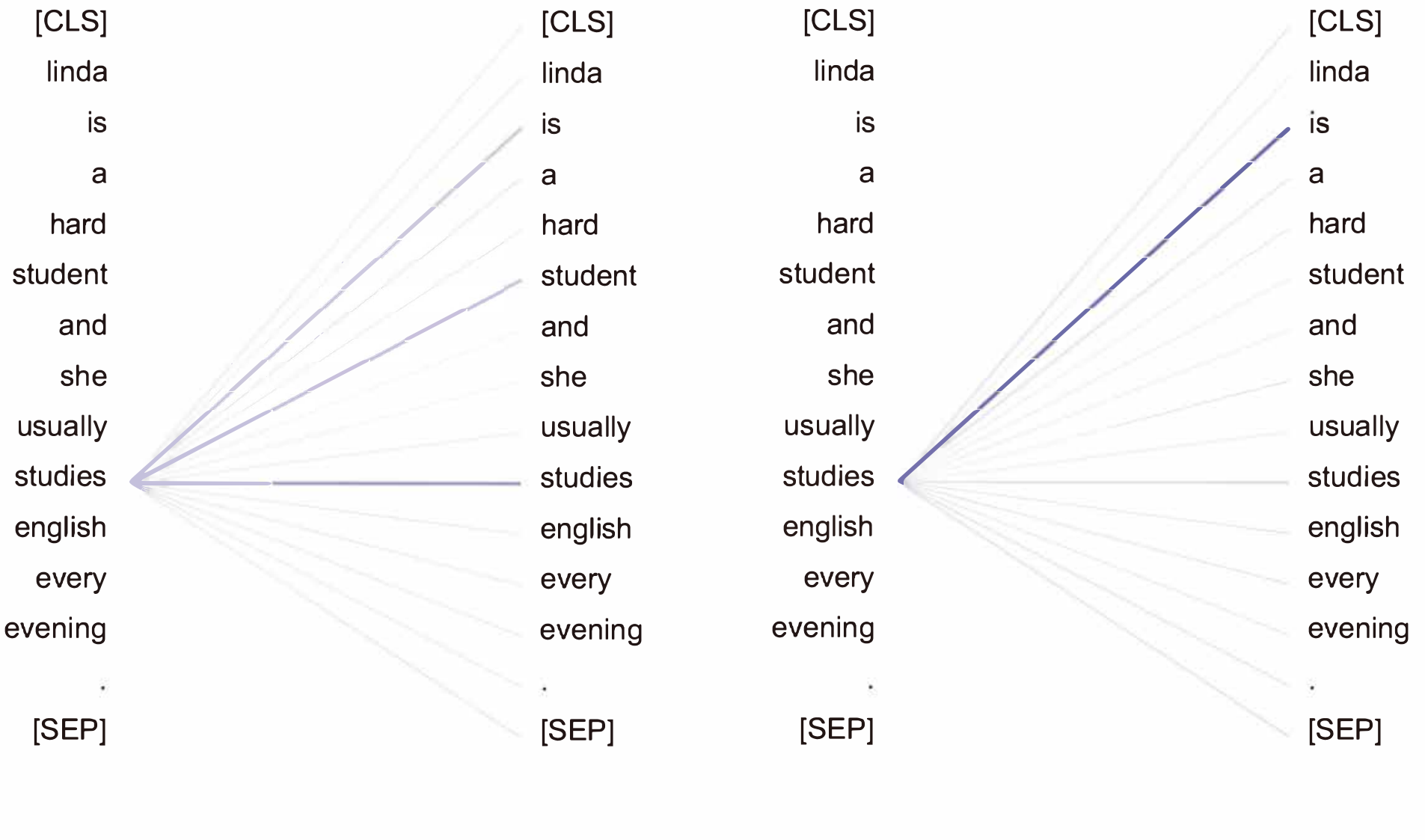}
    \caption{Visualizing the attention weights of layer 3, head 6 in the $\rm BERT_{base\_uncased}$ model (left) and "$\rm BERT_{base\_uncased}$+DSE" model (right). And line thickness reflects the attention weight.}
    \label{fig:visual}
   \vspace{-0.2cm}
\end{figure}

\noindent\textbf{Error Analysis and Visualization on Sentence Completion Task} We analyze the proportion of error samples on different question types. From the results in Fig~\ref{fig:error_ratio}, we can see that the error rates in verbs differentiation, tense consistency and phrase collocation have a significant decrease. It  benefits from syntax information and fine-tuned steps. Furthermore, we use Bertviz, a visualization tool for exploring the attention patterns of the pre-trained BERT model, to visualize the difference of information learned by $\rm BERT_{base\_uncased}$ model and "$\rm BERT_{base\_uncased}$+DSE" model.

\begin{table}[htb]
  \centering
  \vspace{-0.05cm}
  \setlength{\abovecaptionskip}{-0.5cm}
\setlength{\belowcaptionskip}{0.1cm}
  \caption{Relation extraction results on DDI2013 Extraction task. $P$, $R$ and $F_{1}$ represent the metrics of precision, recall and micro f-scores respectively.}
    \begin{tabular}{c|ccc}
    \toprule
    Method        & $P$            & $R$           & $F_{1}$ \\
    \midrule
    \midrule
  Multichannel CNN (Quan et al. \cite{multiCNN})& 75.9        & 62.2        & 70.2 \\
  Hierarchical RNNs  (Zhang et al. \cite{hierarchical_ddi})& 74.1         & 71.8         & 72.9 \\
 One-Stage Model Ensemble (Lim et al. \cite{lim_ddi})& 77.8         & 69.6         & 73.5 \\
    \midrule
    $\rm BERT_{base\_ cased}$ & 72.6        & 66.3        &69.3\\
    $\rm BERT_{base\_cased}$+DSE(concat) &79.6  &65.3  &71.8\\
   $ \rm BioBERT_{base\_cased}$ & 75.5        & 73,2        & 74.4 \\
     $ \rm BioBERT_{base\_cased}$+DSE(concat) & \textbf{77.7} & \textbf{72.4} & \textbf{75.1} \\
    \bottomrule
    \end{tabular}%
  \label{tab:ddi_res}%
   \vspace{-0.4cm}
\end{table}%

\begin{table}[htb]
  \centering
  \vspace{-0.05cm}
  \setlength{\abovecaptionskip}{-0.2cm}
\setlength{\belowcaptionskip}{0.1cm}
  \caption{The result of ablation study}
    \begin{tabular}{c|cc}
    \toprule
     Triple form & ESG dataset(Acc)    &  DDI dataset ($F_{1}$) \\
    \midrule
    \midrule
    ${x_i} = w_{i} \oplus r_{i} \oplus w_{i}^{H}$          & \textbf{91.2}        & \textbf{75.1} \\
    ${x_i} = w{}_i \oplus w_i^H$         & 90.5        & 74.6 \\
    ${x_i} = w{}_i$         & 90.2        & 74.1 \\
    \bottomrule
    \end{tabular}%
  \label{tab:ablation_study}%
 \vspace{-0.5cm}
\end{table}%

\noindent\textbf{Evaluation on DDI task} We compare the metrics of precision, recall, and micro F1 score with recent works of DDI tasks. The results in table~\ref{tab:ddi_res} demonstrate that our model achieved a competitive performance with other state-of-the-art methods. And the syntax expansion information is also beneficial to relation extraction tasks.

\noindent\textbf{Ablation Study} We conduct ablation study on the inputs of dependency fusion layer. To be more specific, we use the DSE model with BERT and change the composition of the triple in equation \ref{eq:concat} to examine the impact of each vector. The experimental result in Table~\ref{tab:ablation_study} shows that dependent relation vector plays a key role in improving the performance of model.

\section{Conclusion}
In this paper, we focused on how to combine syntax information with advanced pre-trained language model. We design a triplet form of dependency tree expansion structure. Further, on the basis of this structure, we propose a novel dependency syntax expansion model. The experimental results both on sentence completion task and relation extraction task prove the effectiveness of our model. And we can draw two conclusions from experimental results. The first is a powerful pre-trained language model can dramatically improve the performance of the system. The second is integrating dependency syntax information with contextualized word embedding can help the model to obtain better sentence representations. In addition, we build a new sentence completion dataset, which is closer to the application scenario. In the future, we will continuously explore the effective method of combing external knowledge with a pre-trained language model and try to apply our model to more downstream tasks.
%
%
%
%

\bibliographystyle{splncs04}
\bibliography{CITE}
\end{document}